\documentclass{Interspeech}

\usepackage{amsmath}
\usepackage{graphicx}
\usepackage{multirow}

\usepackage{chemist}
\usepackage{float}
\usepackage{multicol}
\usepackage{mathtools}
\usepackage{dsfont}

\usepackage{makecell}
\usepackage{colortbl}
\usepackage{rotating}
\usepackage{algorithm}
\usepackage{algpseudocode}
\usepackage{soul}
\usepackage{subcaption}

\interspeechcameraready

\title{Explainable Depression Detection using Masked Hard Instance Mining}

\author[affiliation={1}]{Patawee}{Prakrankamanant}
\author[affiliation={2}]{Shinji}{Watanabe}
\author[affiliation={1}]{Ekapol}{Chuangsuwanich}

\affiliation{Department of Computer Engineering, Faculty of Engineering}{Chulalongkorn University}{Thailand}
\affiliation{Electrical and Computer Engineering}{Carnegie Mellon University}{USA}
\email{6671005621@student.chula.ac.th, shinjiw@ieee.org, ekapol.c@chula.ac.th}

\usepackage{comment}

\begin{document}

\maketitle

\begin{abstract}

This paper addresses the critical need for improved explainability in text-based depression detection. While offering predictive outcomes, current solutions often overlook the understanding of model predictions which can hinder trust in the system.  We propose the use of Masked Hard Instance Mining (MHIM) to enhance the explainability in the depression detection task. MHIM strategically masks attention weights within the model, compelling it to distribute attention across a wider range of salient features. We evaluate MHIM on two datasets representing distinct languages: Thai (Thai-Maywe) and English (DAIC-WOZ).  Our results demonstrate that MHIM significantly improves performance in terms of both prediction accuracy and explainability metrics.

\end{abstract}

\section{Introduction}
Major Depressive Disorder (MDD) is a pervasive mental health condition affecting individuals worldwide \cite{world2017depression}, and the World Health Organization (WHO) estimates it as a leading cause of disability. Despite its widespread impact, many individuals remain unaware of their susceptibility to depression, which often leads to delayed diagnosis and treatment \cite{mathers2006projections}. In recent years, there has been a surge of research that leverages machine-learning (ML) techniques to address the challenges associated with depression detection \cite{burdisso2019text, yalamanchili2020real}. These advanced methodologies have the potential to enhance the diagnostic accuracy and facilitate early intervention.


In recent years, various text-based ML models have been applied to the depression detection task, utilizing both non-deep learning \cite{orabi2018deep} and deep learning approaches \cite{wolohan2018detecting, cong2018xa}. The state-of-the-art model for this task is the Dual Encoder with the attention mechanism \cite{lau2023automatic}. However, \textit{explainability} in this task remains a challenge and underexplored. Some studies have attempted to improve model explainability, such as prompting LLMs to highlight input relevance for prediction \cite{bao2024explainable}. However, these methods often result in lower performance compared to fine-tuning small pre-trained models \cite{bao2024explainable}.




We hypothesize that the attention mechanism in the Dual Encoder struggles with explainability because it is trained on a low-resource dataset \cite{boonvitchaikul2023application, gratch2014distress}. As a result, the attention mechanism overfits to certain features while ignoring other important ones. To address this, we propose masking over-attended features, allowing the model to focus on alternative relevant features. In this study, we view depression detection as a Multi-Instance Learning (MIL) \cite{babenko2008multiple, ilse2018attention} task which treats each part of the dialogue as instances. We introduce a Dual Encoder-based model \cite{lau2023automatic} for depression detection with an enhanced explainability capability using the Masked Hard Instance Mining (MHIM) method \cite{tang2023multiple}. The MHIM method masks features with high attention scores, encouraging the model to consider other significant features. Our experiments on two datasets, Thai-Maywe \cite{boonvitchaikul2023application} and DAIC-WOZ \cite{gratch2014distress}, demonstrate that the MHIM-enhanced model achieves higher performance and improved explainability compared to the dual encoder and LLM-based baselines. Our key contributions are \textbf{(1)} We propose the use of MHIM to the depression detection model based on the principles of MIL to enhance the model’s performance, and \textbf{(2)} The proposed model demonstrates improved explainability within the MIL framework by leveraging attention scores from the attention mechanism, thereby providing interpretable insight into the prediction process.

\section{Background} \label{sec:relate_work}

\subsection{Dual Encoder Model}

\begin{figure*}[!htb]
  \centering
  \centerline{\includegraphics[width=0.8\textwidth]{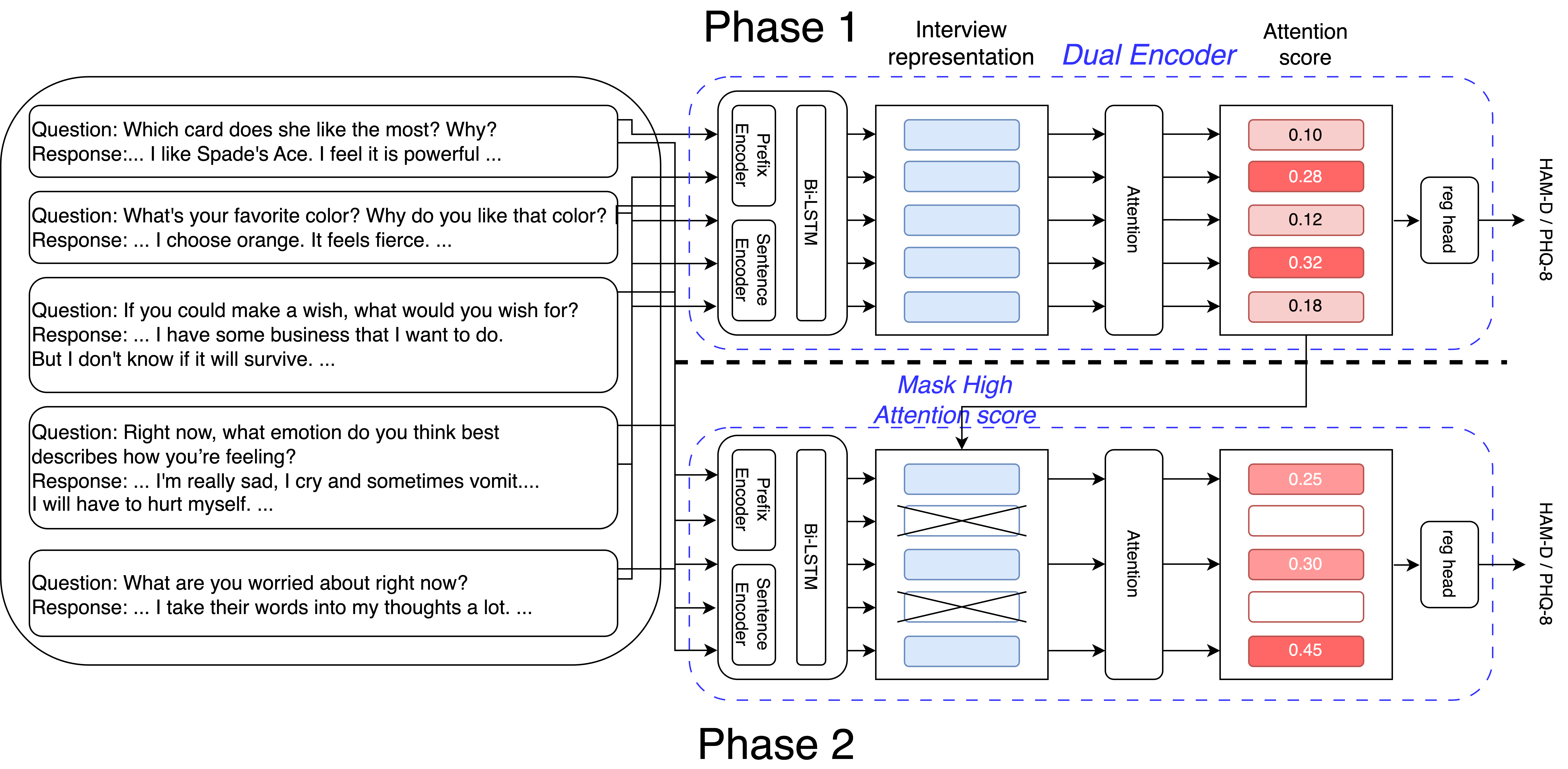}}
  \caption{The overview of our proposed pipeline. The training process consists of two phases: 1 and 2. The first phase follows the standard Dual Encoder training procedure, where each input sentence is encoded and aggregated to form an interview representation. This representation is then processed through an attention layer to predict the depression severity score (HAM-D or PHQ-8). In the second phase, the process remains similar, but the attention scores from the first phase are used as guidance for training using Masked Hard Instance Mining (MHIM) which masks out the sentence embedding which has high attention scores. This discourage the model from focusing on sections with high attention scores, promoting a more balanced learning.
  } \label{fig:sys_pipeline}
\end{figure*}

The state-of-the-art (SOTA) model for text-based depression detection is the the Dual Encoder \cite{lau2023automatic}. As shown in the top half of Figure \ref{fig:sys_pipeline}, the dialogue between the interviewer and the subject is broken down into parts such as question and response pairs. These parts are then encoded using two different encoders, a Prefix Encoder \cite{liu-etal-2022-p} and a Sentence Encoder \cite{cer-etal-2018-universal}. The outputs of these encoders are then aggregated using a Bi-LSTM followed by an attention layer which summarizes the entire conversation for the final output prediction. Notably, the Dual Encoder model, with its attention mechanism, explicitly follows the principles of MIL, making it well-suited for this task.

\subsection{Multi-instance learning}
Multiple Instance Learning (MIL) \cite{ilse2018attention, babenko2008multiple, ramon2000multi} is a method that treats input as a “bag of instances” typically applied in classification tasks. This method is commonly used to address issues related to weakly labeled data \cite{zhou2018brief}. In the weakly labeled setting, the data is represented as a set or group, where individual labels for each instance within the set are not explicitly provided. This approach allows the model to process inputs in the form of sets. In MIL supervised learning for binary classification, the input consists of bags of instances,  $\{X_1, X_2, \ldots, X_n\}$, where each bag $X_i$ contains multiple sub-instances $x_{ij}$, such that $X_i = \{x_{i1}, x_{i2}, \ldots, x_{im}\}$, with $x_{ij} \in \mathcal{X} = \mathbb{R}^d$. The label for each bag is denoted as $y_i \in \mathcal{Y} = \{0, 1\}$. If the labels of individual instances within a bag $y_{ij}$ where $y_{ij} \in \mathcal{Y}$, are known, the relationship between the bag label $y_i$ and the instance labels $y_{ij}$ can be expressed by Equation \ref{eq:mil_framing_problem}.

\begin{equation}
    y_i = \begin{cases}
            1 &\text{if $\exists j$ s.t. $y_{ij} = 1$}\\
            0 &\text{otherwise}
            \end{cases} \label{eq:mil_framing_problem}
\end{equation}

To address the MIL problem, an aggregation function is required to combine individual input representations into a unified representation. Attention mechanisms are particularly popular due to their effectiveness in modeling complex relationships.


In our context, for each subject $i$, an entire interview session ($X_i$) comprises of multiple responses or conversation turns ($x_{ij}$), where each has an associated level of severity ($y_{ij}$). We only observe $y_i$, a single positive value, indicating severity level as diagnosed by the human expert.

\subsection{Masked Hard Instance Mining (MHIM)}
MHIM \cite{tang2023multiple} is a data augmentation technique designed for MIL models that use attention mechanisms \cite{ilse2018attention, shao2021transmil} to aggregate features from a bag of instances \cite{babenko2008multiple}. This method creates ``hard'' positive or negative samples by masking portions of the input within a bag, thereby effectively augments challenging samples into training data. It encourage the model to focus on other important instances, preventing it from overly relying on a few highly attended instances.

\subsection{Explainability through Attention Mechanism}
Since its introduction, attention mechanism \cite{DBLP:journals/corr/BahdanauCB14} proved to be a long-standing vital component in every aspect of modern deep learning. In addition to superior performance and generalization ability, its intuitiveness and adjacency to human attention also provide a nice explainability touch \cite{clark-etal-2019-bert, sen-etal-2020-human, 9671639}. By analyzing the attention weights, we can identify the important features from inputs.

\begin{table*}[!htb]
\caption{Performance on the Thai-Maywe dataset. Best performers in each metric are highlighted in \textbf{Bold}. We report the mean and standard deviation across five folds.} \label{tab:main_result}
\centering
\begin{tabular}{|l|ccc|ccc|ccc|}
\hline
 & \multicolumn{3}{c|}{HAM-D 1} & \multicolumn{3}{c|}{Overall HAM-D score} \\ \cline{2-7} 
 & RMSE & MAE & Entropy & RMSE & MAE & Entropy \\ \hline
Dual Encoder \cite{lau2023automatic} &  $0.54 \pm 0.28$  & $0.37 \pm 0.19$ & $2.406$ & $3.54 \pm 1.00$ & \textbf{2.47} $\pm$ \textbf{0.76} & $2.158$ \\

Dual Encoder w/ MHIM (ours) & \textbf{0.48} $\pm$ \textbf{0.28} & \textbf{0.33} $\pm$ \textbf{0.19} & $2.496$ & \textbf{3.15} $\pm$ \textbf{0.83} & $2.52 \pm 0.60$ & $2.819$ \\

Single Encoder \cite{milintsevich2023towards} & $0.58 \pm 0.21$ & $0.44 \pm 0.20$ & $2.722$ & $3.56 \pm 1.08$ & $2.73\pm 0.94$ & $2.725$ \\

Single Encoder w/ MHIM (ours) & $0.57 \pm 0.21$ & $0.40\pm 0.18$ & $2.735$ & $3.40 \pm 0.32$& $2.56 \pm 0.81$ & $2.957$ \\

LLM &  $1.08 \pm0.29 $ &  $0.65 \pm 0.11$ & - & $6.34 \pm 0.15$ & $4.42 \pm 1.02$ & - \\ \hline
\end{tabular}
\end{table*}

\section{Methods} \label{sec:method} 

In the Dual Encoder, the input sentences are fed directly into the model for depression score prediction. We find that this procedure often suffer from overfitting, especially on low-data regime. Our investigation indicates that attention mechanism in the Dual Encoder tend to concentrate to only a small portion of the input features, exhibiting low entropy attending pattern. Additionally, in the depression detection task, the model might attend to only a couple of sentences that exhibit strong indicators. For example, the model might focus on just the mentioning of suicide, even though there are sentences about troubled sleeping which might be useful information for the care provider. Thus, for explainability purposes, we might want the attention weights to cover all of the factors. We believe that by spreading out the attention weights, more diverse information aggregation is encouraged, providing better prediction performance and explainability at the same time.

Under our hypothesis, we propose the use of Masked Hard Instance Mining (MHIM) in the attention layer of the Dual Encoder. As shown in Figure \ref{fig:sys_pipeline}, our method consists of 2 phases. \textbf{(Phase-1)} We follow the \cite{lau2023automatic} training scheme and train a Dual Encoder to predict the HAM-D or PHD-Q score. This model will subsequently act as the Mask-Doner for our phase-2's model, in which doner's attention weights will be used to determine attention mask for the receiver. The masks comes from two sets: top scoring masks and random masks. $t$ top scoring masks are randomly chosen from the top $r$ attention weights ($t \le r$). For the bottom attention weights, we randomly select $b$ masks, for a total of $N = t + b$. The bottom attention weights are masked to encourage diversification which helps with generalization. \textbf{(Phase-2)} Next, another Dual Encoder is trained similar to the doner, but with an additional of attention masking, using masks received from the doner. This effectively blinds the receiver from attending to the doner's top attentive tokens, encouraging diverse attention pattern in the receiver.

Afterwards, we perform inference using the original Dual Encoder's scheme, feeding input sequences directly into the receiver, bypassing the entire masking operation.

\section{Experiment Setup} \label{sec:exp}

\subsection{Datasets}

To benchmark the performance of our method, we have chosen two datasets for benchmarking: the DAIC-WOZ \cite{gratch2014distress} and the Thai-Maywe dataset \cite{boonvitchaikul2023application}. The DAIC-WOZ dataset is a set of 189 clinical interviews conducted in English, utilizing the PHQ-8 depression screening questionnaire \cite{kroenke2009phq}, which includes eight items. For this study, we focus specifically on the second item, which assesses feelings of depression and helplessness (143 interviews). The Thai-Maywe dataset is a dataset of 84 patients undergoing psychiatric evaluations for depression obtained through a mobile application and assessed using the Hamilton Depression Rating Scale (HAM-D or HDRS)  \cite{hamilton1986hamilton, regier2013dsm}, which consists of 17 symptom-related items (HAM-D 1-17). The 17 items are combined to give an overall score. This dataset also provides Importance Sentence Labels (ISL). A human expert gives a binary label to each answer in the interview indicating whether the answer is relevant to the HAM-D item. In our work, we only focus on HAM-D 1 (Depressed Mood) since ISL labels are only available for this item.  In both datasets, the interviews were machine transcribed into text which was used as is in our experiments.

For the DAIC-WOZ dataset, since labels are not provided in the official test set, we split one-fifth of the training set to create a new validation set, which is used for reporting. For the Thai-Maywe dataset, we use five-fold cross-validation with a 3:1:1 train-validation-test split. 


\subsection{Baselines}

For the DAIC-WOZ dataset, we followed the prior work \cite{lau2023automatic} and employed RoBERTa-base \cite{yinhan2019roberta} and all-mpnet-base-v2 \cite{song2020mpnet} as the pretrained models for the Prefix Encoder and Sentence Encoder, respectively. On the other hand, we used WangchanBERTa \cite{lowphansirikul2021wangchanberta} and ConGen \cite{limkonchotiwat2022congen} as the Prefix Encoder and Sentence Encoder, respectively, for the Thai-Maywe dataset as these models are more suitable for the Thai language.

The Dual Encoder was trained using Mean Squared Error (MSE) loss with a learning rate of $3 \times 10^{-5}$, 10\% learning rate warm up, followed by a linear decay, both starting and ending at $1 \times 10^{-7}$. We trained for 200 epochs using the AdamW optimizer. The hyperparameters $t$ and $b$ were selected based on the validation set, and set $r=2t$.



In addition to the Dual Encoder, we also assessed the effectiveness of our methodology on the Single Encoder model \cite{milintsevich2023towards}, which shared the same architecture as the Dual Encoder but utilized only the Sentence Encoder. Both models were implemented using the official GitHub repository\footnote{\url{https://github.com/clintonlau/dual-encoder-model}}, and the Single Encoder followed the same training configuration as the Dual Encoder. We also compared the results with LLM-based prompting using Gemini-2.0-Flash-001, following the prompting methodology outlined in \cite{bao2024explainable} by generating output in XML format. To be comparable with supervised models, all of the training data are provided in the prompt as few-shot examples.

\subsection{Evaluation Metrics}

We evaluated the performance of our method on two aspects: prediction performance, and explainability. For prediction performance, we directly used RMSE and MAE as evaluation metrics. For explainability, we aimed to measure how well the model identified sections with high relevance to HAM-D. To this end, we employed Recall@k as the evaluation metric, as defined in Equation \ref{eq:recall_k}.
\begin{equation} \text{Recall}@\text{k} = \frac{\sum_{i}^{n} \mathds{1}_A(a_i,k) \cdot y_{i}}{\sum_{i}^{n} y_{i}} \label{eq:recall_k} \end{equation}

where $n$ represents the number of sentences in the sample, $A$ denotes the attention scores, $a_i$ is the attention score of the $i^{th}$ sentence, and $y_i$ is the ISL score (0 or 1) for the $i^{th}$ sentence. The indicator function $\mathds{1}_A(a_i,k)$ returns 1 if $a_i$ is ranked within the top-k attention scores, and 0 otherwise.



\section{Result \& Discussion} \label{sec:exp}

\subsection{Model Performance on the Thai-Maywe dataset} \label{ssec:main_result}


Table \ref{tab:main_result} shows the prediction performance on the Thai-Maywe dataset of our method compared to other baselines. Both the Dual Encoder and Single Encoder models achieve improved performance by using the MHIM method, as measured by RMSE and MAE. Moreover, the MHIM properly encourages the attention mechanism to spread out the attention scores. We calculated the entropy of the attention weights in the four models. The entropy increases when the MHIM is applied.

\subsection{Model Explainability on the Thai-Maywe dataset} \label{ssec:explain}

\begin{table}[!ht]
\centering
\caption{Recall@k results of the Dual Encoder model with and without the MHIM method. For the dual encoder model, important sentences are selected by ranking based on the attention weights. In the LLM-based method, the LLM only selects a limited number of sentences as important sentences. In such cases, we use * to indicate that the number of sentences selected does not reach $k\%$.}
\label{tab:hand_recall}
\begin{tabular}{|c|ccccc|}
\hline
 & \multicolumn{5}{c|}{Recall@k} \\ \cline{2-6} 
 $k=$ & $10\%$& $20\% $ & $50\%$& $80\%$& $90\%$ \\ \hline
 Dual Encoder  & $0.13$ &  $0.21$ & $0.52$ & $0.85$ &  $0.93$\\
 +MHIM & $0.17$ & $0.27$ & \textbf{0.62} & \textbf{0.90} & \textbf{0.95} \\ 
 LLM & \textbf{0.22} & \textbf{0.28} & $0.35$ & $0.36^{*}$ & $0.36^{*}$ \\ \hline
\end{tabular}
\end{table}

In this section, we aim to investigate whether the 2-phase training scheme can enhance the models' explainability. We compare the top attended features against the ISL's choices. Table \ref{tab:hand_recall} shows the Recall@k on HAM-D 1 for the different models. The dual encoder model with MHIM performs better than the one without. The LLM baseline performs well only on the top sentences (lower $k$), but struggles when prompted to provide a longer list of explanations.

For the overall HAM-D, since there are no ISL labels. Instead, we performed a kind of sensitivity analysis on how the model prediction will change if different sentences were dropped from the input. If important sentences are dropped, it should have a high impact on the model's performance.

\begin{figure}[!ht]
\centering
\includegraphics[width=0.70\linewidth]{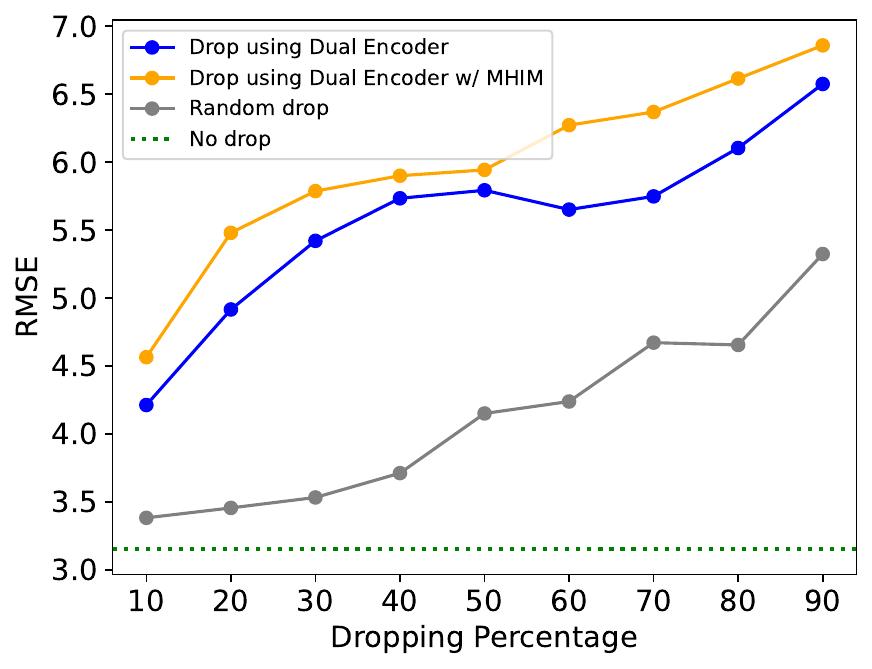}
\caption{Model prediction results on the Thai-Maywe dataset (overall HAM-D) when the top-k features are dropped using different dropping criterion. Removing features selected by our method has the biggest effect on the prediction score.}
\label{fig:attn_sel_sum_HAMD}
\end{figure}

Figure \ref{fig:attn_sel_sum_HAMD} shows the effect of dropping the important sentences according to the top attention scores. The figure compares dropping sentences using the attention scores from the Dual Encoder baseline, the Dual Encoder with MHIM, and random dropping. The plot demonstrates that our technique is more effective at identifying important sentences, as shown by the upward RMSE shift compared to other methods.

Table \ref{tab:cherry_pick_02} shows the top 3 sentences selected by different models. As mentioned in Section \ref{sec:method}, the Dual Encoder tends to provide bad explainability candidates via the attention scores. However, with the  MHIM in the two-phase training, more meaningful sentences can be identified by both architectures. 

Although the overall results look promising, our model is prone to cases where a single high-ISL sentence may lead to a false positive. We believe that this might be caused by the MIL framing which we plan to remedy in our future work.

\begin{table}[h]
\caption{The 3 most important sentences by each model, presented as translated excerpts from a HAM-D 1 interview. \textbf{Bold} indicates that the sentence is relevant according to the ISL.} 
\label{tab:cherry_pick_02}
\centering
\begin{tabular}{l}
\hline
\multicolumn{1}{c}{Dual Encoder} \\ \hline \hline

\makecell[l]{... I eat normally. ...}  \\
\makecell[l]{... Using this application is a little difficult. ...} \\ 
\makecell[l]{... \textbf{I had a terrible feeling this past weekend.} ...} \\ \hline



\multicolumn{1}{c}{Dual Encoder with MHIM} \\ \hline \hline

\makecell[l]{... \textbf{Lately, I don't really feel like meeting anyone.} ...} \\
\makecell[l]{... \textbf{I've been feeling sad from time to time lately.} ... } \\
\makecell[l]{... {I try to write to release stress.} ... } \\ 
\hline

\multicolumn{1}{c}{Single Encoder} \\ \hline \hline

\makecell[l]{... \textbf{Lately, I don't really feel like meeting anyone.} ... } \\
\makecell[l]{... I try to write to release stress. ...} \\
\makecell[l]{... I don't think about hurting myself anymore. ...} \\ 
\hline

\multicolumn{1}{c}{Single Encoder with MHIM} \\ \hline \hline

\makecell[l]{... \textbf{Lately I've been worried and unable to sleep.} ...} \\ 
\makecell[l]{... \textbf{I can't sleep and am stressed about it.} ... } \\ 
\makecell[l]{... \textbf{Lately, I don't really feel like meeting anyone.} ... } \\
\hline

\multicolumn{1}{c}{LLM} \\ \hline \hline
\makecell[l]{... \textbf{I'm a bit worried about money.} ...} \\
\makecell[l]{... {I go out to eat and travel to relieve stress.} ...} \\
\makecell[l]{... \textbf{I've been feeling sad from time to time lately.} ...} \\ 
\hline
\end{tabular} 
\end{table}

\subsection{Results on the DAIC-WOZ dataset}

We employ our technique on the DAIC-WOZ dataset, specifically for PHQ-8 item 2 (Depressed and Hopeless). Using the same Dual Encoder architecture, the prediction performance with and without the MHIM technique are comparable. The RMSE decreases from 0.733 to 0.717 when MHIM is applied.

\begin{figure}[!ht]
\centering
\includegraphics[width=0.70\linewidth]{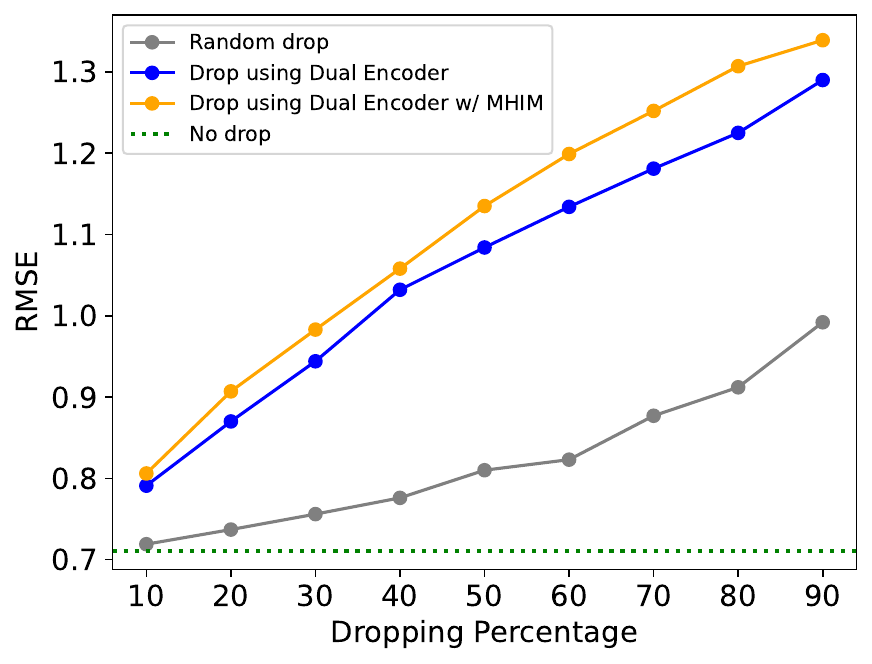}
\caption{Prediction results on the DAIC-WOZ dataset when the top-k features are dropped using different dropping criterion.}
\label{fig:attn_sel_phq_depress} 
\end{figure}

For explainability evaluation, we apply the same analysis used previously, dropping features based on top attention values. As shown in Figure \ref{fig:attn_sel_phq_depress}, the results are similar to those observed in the overall HAM-D score. Thus, our technique is effective in improving explainability while maintaining prediction performance.

\section{Conclusion} \label{sec:con}


We proposed a 2-phase attention-based depression detection model training pipeline that leveraged the Masked Hard Instance Mining (MHIM) technique and the Multi-instance Learning (MIL) framework. The pipeline was specifically designed to promote diversified attending pattern, thereby less prone to overfitting in low-data settings. We found that the model trained with the MHIM method not only improved performance but also enhanced explainability through the attention scores within the model in both the Thai-Maywe and DAIC-WOZ datasets. Further improvements such as better utilization of the doner's attention weights, or method less stringent than complete masking are left for the future work.

\section{Acknowledgements}
This research project is supported by the Second Century Fund (C2F), Chulalongkorn University.

\bibliographystyle{IEEEtran}
\bibliography{mybib}


\end{document}